\documentclass[letterpaper, 10 pt, conference]{ieeeconf}
\IEEEoverridecommandlockouts    

\usepackage{cite}
\usepackage{amsmath,amssymb,amsfonts}
\usepackage{algorithmic}
\usepackage{graphicx}
\usepackage{textcomp}
\usepackage{xcolor}
\usepackage{tikz} 
\usetikzlibrary{positioning, arrows.meta}
\usepackage{url}
\usepackage{subcaption}
\usepackage{censor}
\usepackage{booktabs}

\usepackage{soul}  
\sethlcolor{yellow} 

\def\BibTeX{{\rm B\kern-.05em{\sc i\kern-.025em b}\kern-.08em
    T\kern-.1667em\lower.7ex\hbox{E}\kern-.125emX}}

\newcommand\copyrighttext{%
\footnotesize \copyright 2026 IEEE. Personal use of this material is permitted. Permission from IEEE must be obtained for all other uses, in any current or future media, including reprinting/republishing this material for advertising or promotional purposes, creating new collective works, for resale or
redistribution to servers or lists, or reuse of any copyrighted component of this work in other works.}
\newcommand\copyrightnotice{%
\begin{tikzpicture}[remember picture,overlay]
\node[anchor=north,yshift=-1cm] at (current page.north) {\parbox{\dimexpr\textwidth-\fboxsep-\fboxrule\relax}{\centering \copyrighttext}};
\end{tikzpicture}%
}
    
\begin{document}

\title{\textbf{High-Fidelity Multi-Body Simulator for Autonomous Racing}
}

\author{
{Nicola Musiu$^{1}$, Francesco Iacovacci$^{1}$, Fausto Lupo$^{1}$, Matteo Pini$^{1}$, Giovanni Scapicchi$^{1}$,}\\ 
{Francesco Moretti$^{1}$, Eugenio Mascaro$^{1}$, Pietro Musso$^{1}$, Ayoub Raji$^{1}$, Marko Bertogna$^{1}$,}\\
{Vincenzo Maria Arricale$^{2}$, Angelo Lo Sapio$^{3}$, Alessandro Piccarelli$^{4}$, Garron Fish$^{4}$}%
\thanks{$^{1}$University of Modena and Reggio Emilia, Modena, Italy.
{\tt\small nicola.musiu@unimore.it}}%
\thanks{$^{2}$Megaride S.r.l., Naples, Italy}%
\thanks{$^{3}$University of Naples Federico II, Naples, Italy}%
\thanks{$^{4}$Claytex, Edmund House, Rugby Road, Leamington Spa, UK}}%

\maketitle
\copyrightnotice

\begin{abstract}

We present a custom high-fidelity vehicle dynamics simulation 
environment for testing and validation of Autonomous Racing software.
The digital twin of the autonomous vehicle is developed in Dymola, using racecar dynamics modeling libraries to build a complete 
multi-body model. A 3D road surface, including elevation profiles and curbs, is implemented using the Curved Regular Grid (CRG) standard.
The model is exported from Dymola as a Functional Mock-up Unit 
(FMU) and integrated into a custom software-in-the-loop simulator, where 
communication interfaces with the autonomous racing stack were developed 
in C++.
A calibration procedure based on experimental data is also presented, 
along with a validation study to further support the quality of the 
proposed framework.
The simulator runs in real time on a portable computer and provides 
reliable ground truth for algorithms validation prior to real-world 
deployment.


\end{abstract}


\section{Introduction}
Autonomous Racing has emerged as a research laboratory for autonomous driving hardware and algorithms, thanks to the unique challenges of motorsport. 
However, 
the cost and risk associated with experimental testing are extremely high.
Consequently, track time is often limited 
and is reserved exclusively for fine-tuning activities.
Furthermore, especially during the early stages of a project, vehicles 
may not be physically available or fully testable. As a result, 
systematic on-track data collection and 
parameter identification cannot be performed, creating a critical gap 
between algorithm development and experimental validation.


A tool capable of virtually reproducing single maneuvers or 
entire track sessions with both high efficiency and high fidelity, therefore, becomes essential.

Motivated by this need, we developed a custom vehicle dynamics simulation 
environment. The proposed framework enables real-time testing and validation of an Autonomous Racing stack, providing reliable ground truth for algorithm assessment prior to real-world deployment and significantly accelerating the development process.

\begin{figure}[t]
  \centering
    \includegraphics[width=1.\linewidth, trim=0 0 20 20, clip]
    {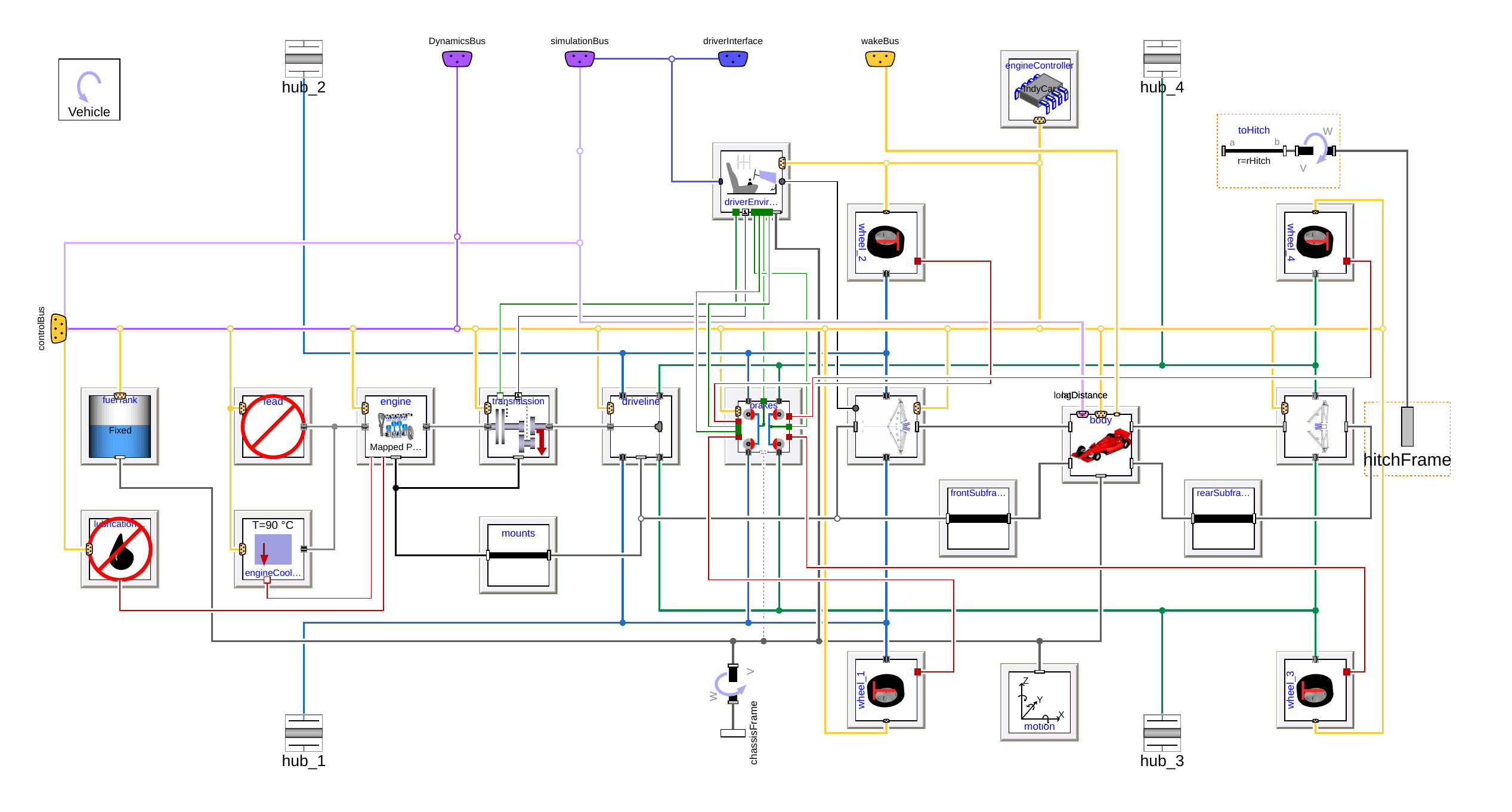}
\caption{Top-level block diagram of the Dymola race car model, showing all 
main components and their connections.}
\label{fig:fmu_vehicle}
\end{figure}

The digital twin of the autonomous vehicle is developed in Dymola \cite{dym_1}, a Modelica-based modeling and simulation environment widely adopted in industry for multi-domain physical systems.
A key advantage is the availability of highly specialized libraries for racecar dynamics modeling, such as VeSyMA and Motorsports \cite{vesyma}, which provide template vehicle architectures adaptable to specific requirements (Figure \ref{fig:fmu_vehicle}). 
These libraries accelerate model development and enable the assembly of accurate, fully parameterizable models with minimal low-level implementation effort.

This environment is fully compatible with the OpenCRG format \cite{OpenCRG}, an open standard for high-resolution 3D road surface description. 
In addition, Dymola natively supports the Functional Mock-up Unit (FMU) 
standard \cite{FMI_2_0}, which addresses the challenge of co-simulating dynamic models developed with heterogeneous software tools by defining an open, tool-independent interface for integration within unified simulation 
frameworks.

The FMU-based vehicle model and the CRG road representation are 
integrated into a custom C++ software-in-the-loop simulation environment 
using the FMI4cpp library \cite{FMI4cpp}, thereby creating the complete simulation framework.

%
%
%
%

Section~\ref{sec:soa_contrib} reviews the state of the art in simulation tools for Autonomous Racing and highlights the contribution of this work.
Section~\ref{sec:workflow} describes the development workflow.
Section~\ref{sec:modeling} details the main components of the proposed framework, including vehicle and road models, FMU packaging, and related stability and computational efficiency aspects.
Section~\ref{sec:calib} briefly presents the tire model calibration procedure.
Section~\ref{sec:sim_env} introduces the final simulation environment and its integration interface.
Section~\ref{sec:results} reports simulation accuracy by comparison with experimental data.
Conclusions and future work are presented in Section~\ref{sec:conclusion}.

\section{State of the Art and Contribution}\label{sec:soa_contrib}

The literature on simulators for Autonomous Racing varies from solutions built on Unity or CARLA environments \cite{scapicchi2024ar,purdue,brunner2025mathcal}, leveraging their modularity and extensibility, particularly for the sensors configuration, to platforms specifically focused on training Machine Learning agents \cite{learntorace,wymann2000torcs}. 
In these works, the dynamics of the vehicle remain relatively limited, with some degree of potential integration with external custom models.

More recently, arcade-style gaming environments
are increasingly being adopted for autonomous agent testing. 
For instance, recent works have employed Gran Turismo (GT) 
for Reinforcement Learning (RL) agent training, demonstrating that autonomous agents can achieve superhuman lap times \cite{GT_1}. 
However, when such approaches are evaluated in more realistic and less 
permissive environments, they do not consistently outperform human drivers 
under all conditions, as shown in~\cite{Adrian}, where autonomous agents
are evaluated using Assetto Corsa (AC).
This observation highlights the importance of high-fidelity vehicle 
dynamics simulation in Autonomous Racing, where performance 
evaluation requires simulation environments capable of accurately 
predicting real-world outcomes \cite{zimon}.

By contrast, vehicle-dynamics-oriented simulators are typically developed 
for human driver-in-the-loop applications \cite{dym_2,dym_3}, prioritizing 
accurate modeling of vehicle’s physical behavior and are therefore 
rarely adopted in Autonomous Racing. A satisfactory level of accuracy is achieved in \cite{zimon}, where a simplified multi-body model written in C++ is presented and publicly released. Nevertheless, its comparison with real data is limited to the analysis of metrics related to the autonomous controller (lateral error) without any specific demonstration of dynamics fitting. Furthermore, the closeness to the vehicle limits in the data used is not mentioned.


Motivated by these considerations, we developed a custom simulation framework that provides full control over the modeling assumptions, complete tuning and parameterization based on experimental data, and that, compared to black-box physics engines, fosters a deeper understanding of the system dynamics and supports the development of model-based control algorithms, where physics-based approaches are commonly adopted \cite{Musiu}.
In particular, we present the employed complete workflow, providing practical guidelines for the research community. The paper includes a detailed overview of the modeling techniques and tools employed to achieve a realistic yet capable of real-time performance simulation framework for Autonomous Racing.
Other contributions of the work are:
\begin{itemize}
    \item A fully customizable and parameterizable high-fidelity, vehicle-dynamics-oriented simulator tailored to Autonomous Racing applications.
    \item Real-time capability with low implementation effort and reduced 
    costs, enabled by the adoption of widely available open-source tools;
    \item Compliance with industrial standards commonly adopted in the 
    automotive industry (e.g. FMU and CRG), thus promoting flexibility and scalability.
\end{itemize}

\section{Development of the Simulation Framework}\label{sec:workflow}

The Dymola model was initially populated with all parameters required 
by the VeSyMA and Motorsport libraries, most of which were 
provided by the vehicle manufacturer 
(chassis, tires, suspensions, powertrain and aerodynamics).
This enabled a first physics-based 
approximation of the vehicle behavior prior to on-track validation, 
allowing the model to be used for in-depth vehicle-dynamics-oriented 
simulations.

Dymola simulations were performed to generate maneuvers that are 
difficult to reproduce experimentally, particularly when on-track 
testing is unsafe or impractical, or when the autonomous vehicle is not 
physically available. This approach provided high-quality reference data 
for the calibration of simplified models for path planning and control, including single-track 
representations, where accurate identification of axle characteristics is essential \cite{Musiu}.

In addition, the simulations on Dymola were exploited to derive simplified model 
subcomponents, such as the force distribution of a locked differential, 
a reduced roll-axis lateral load transfer model,
and the contribution of road banking to vertical tire loads.
These analyses support the development of enhanced control-oriented models that improve planning and control performance \cite{Raji_2023}.

The autonomous software was initially validated on a Unity-based simulator \cite{er1_1}, providing a trade-off between computational efficiency and vehicle dynamics fidelity.
However, the fidelity and customizability of this setup were not comparable 
with the multi-body model developed in Dymola; therefore, we aimed to 
leverage the Dymola model as a high-fidelity physics engine.
Following recent improvements aimed at enhancing the computational 
efficiency of our high-fidelity model (detailed in Section \ref{sub_sec:exec_time}), 
it became possible to export it as a Functional Mock-up Unit (FMU) and 
integrate it into a custom software-in-the-loop simulator. 
This step represents the final stage of the development, 
enabling real-time testing of the complete autonomous software, as detailed 
in Section~\ref{sec:sim_env}.

From the road modeling perspective, the full-stack simulations were initially run on a flat surface. 
This approximation proved too coarse, leading first to the introduction of a variable banking angle 
along the track. More recently, the road has been modeled using high-resolution 3D surfaces in CRG format, leveraging the OpenCRG framework \cite{OpenCRG} for accurate representation of complex geometries. Further details are provided in Section~\ref{subsec:road_model}.

\section{Modeling}\label{sec:modeling}


\subsection{Vehicle Model}

The Dymola vehicle model is organized as a set of interconnected 
blocks
, each representing a different vehicle subsystem contributing to the overall vehicle dynamics.
In their standard configuration, the adopted Motorsport libraries are 
sufficient to model a complete race vehicle with high fidelity. 
However, selected components have been extended to meet specific requirements of the application:

\begin{itemize}

\item \textbf{Tires}:  
The force–slip relationship is modeled using a Pacejka Magic Formula 6.2.  
A custom thermal model captures variations in grip and stiffness 
as a function of temperature, as detailed in the following paragraphs.

\item \textbf{Brakes}:  
The original block has been extended to support four independent brake commands. The standard brake thermal model has been further enhanced to account for a varying pad–disk friction coefficient.

\item \textbf{Driveline}:  
A centrifugal clutch progressively engages from 1400 rpm to full 
engagement at 1800 rpm.

\item \textbf{Differential}:  
Implements both a fully locked configuration and a 
friction-based limited slip differential.

\item \textbf{Suspensions}:
 Standard models were adapted to represent the 
geometry of the actual autonomous vehicle.

\item \textbf{Aerodynamics}:  
Slipstream effects from nearby vehicles reduce drag and downforce. The opponent’s position $(x,y)$ is used to query a lookup table that returns scaling coefficients.

\item \textbf{Actuators dynamics}:  
Are modeled through transfer functions identified 
from on-track data. These models are implemented outside the Dymola environment.

\end{itemize}

\noindent
A more detailed view of some components is given.

\subsubsection{Brake Thermal model}

The braking system features a custom two-node 
lumped-parameter thermal model coupled with a data-driven 
brake pad-disk friction map.
The mass of the disk is discretized into two distinct thermal nodes:

\begin{itemize}
    \item \textbf{Surface Node:} Represents the outer friction ring with low thermal inertia, 
    rapidly responding to braking heat flux.
    \item \textbf{Core Node:} Represents the internal bulk and ventilation vanes acting as 
    thermal capacity, exchanging heat with the surface via conduction and 
    dissipating it through convection.
\end{itemize}

\noindent
The temperature evolution is determined by the energy balance of heat entering, 
stored, and leaving the system:
\begin{equation}\label{eq:brake_thermal}
    C_{\text{surf}} \cdot \dot{T}_{\text{surf}} = \dot{Q}_{\text{in}} - \dot{Q}_{\text{conv}} - \dot{Q}_{\text{rad}} - \dot{Q}_{\text{cond}},
\end{equation}
\noindent
where $C_{\text{surf}}$ is the heat capacity of the surface node, which is 
temperature-dependent, and $T_{\text{surf}}$ is the disk surface temperature.
The model accounts for three heat transfer mechanisms:

\begin{itemize}
    \item \textbf{Conduction ($Q_{\text{cond}}$):} Heat exchange between Surface and Core nodes through material thermal conductance.
    \item \textbf{Convection ($Q_{\text{conv}}$):} Heat dissipation to ambient air from external faces and ventilation channels; the heat transfer coefficients are computed from disc angular velocity and vehicle speed using Nusselt correlations.
    \item \textbf{Radiation ($Q_{\text{rad}}$):} Radiative losses modeled via the 
    Stefan-Boltzmann law, dominant at high temperatures typical of carbon discs.
\end{itemize}

\noindent
The braking torque and the heat flux entering the system (\textbf{$Q_{\text{in}}$}) are governed by the pad-disk friction coefficient ($\mu_d$). Unlike standard models, our implementation retrieves $\mu_d$ dynamically from an experimental friction map, as a function of the pad-disk contact pressure and the disk surface temperature.
The generated friction power is scaled by a heat partitioning factor to determine the effective heat input for the thermal model in Eq.\eqref{eq:brake_thermal}.

\subsubsection{Tire Thermal model}

To capture the temperature-dependent behavior of tires, the peak force and the cornering stiffness of the Pacejka Magic Formula are modeled as functions of the tread temperature.
For the lateral characteristics:
\begin{equation}\label{eq:dym_temp}
\begin{aligned}
  D_{y}(T) &= D_{y} \Big(1 + \frac{\delta \mu_y}{\delta T} (T-T_{ref})  \Big), \\
  K_{y}(T) &= K_{y} \Big(1 + \frac{\delta C_y}{\delta T} (T-T_{ref})  \Big). \\
\end{aligned}
\end{equation}
\noindent
where $T$ is the tread temperature and $T_{ref}$ is the reference ambient 
temperature. The variations of the peak friction coefficient $\mu_y$ and of
the cornering stiffness $C_y$ with respect to temperature are
experimentally identified.
In this work, a simplified piecewise approximation is used: grip increases during warm-up, remains constant within a thermal window, and decreases when overheating occurs.
Cornering stiffness decreases monotonically with increasing temperature.
The same reasoning is applied to the longitudinal tire characteristics.
Heat generation is computed internally using tire forces and slip.

Eqs.\eqref{eq:brake_thermal}-\eqref{eq:dym_temp} have been implemented as a custom blocks 
within the existing Motorsport tire and brake models. These integrate thermal 
components from the Modelica standard library such as heat ports connecting lumped 
masses to their thermal environment (air, ground, internal layers), and dedicated 
elements model the corresponding heat transfer.

This integration results in a fully coupled thermo-mechanical model, capable of capturing performance change induced by temperature variations.
Thermal models can be easily disabled, providing a 
more deterministic simulation.

\subsubsection{Actuation models}
Steering and brake actuation include delays and friction effects.
A delay in the throttle command is also introduced to approximate the turbo-lag effect in a simplified manner, acknowledging that the full turbocharger dynamics cannot be represented by a  pure time delay.

The actuators were modeled using a linear transfer function following an ARX structure (autoregressive with exogenous input). The model parameters were estimated from input-output experimental data using the MATLAB System Identification Toolbox. Despite its simple structure, the ARX model captures enough information to represent the actuator-relevant dynamics, of which the dead time delay is the most significant. 

\begin{figure}[b]
    \centering
    \begin{subfigure}{0.155\textwidth}
        \centering
        \includegraphics[width=\linewidth, trim=0 0 400 0, clip]{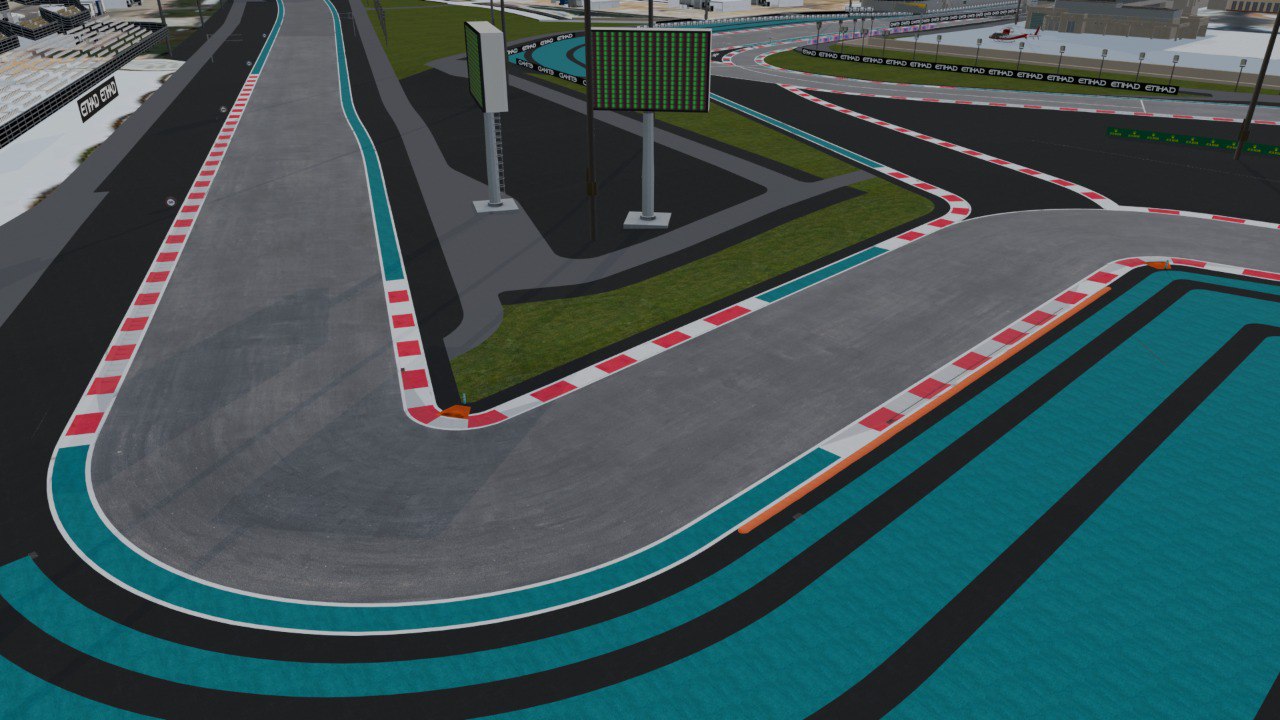}
        \caption{}
        \label{fig:sub1_crg}
    \end{subfigure}
    \hfill
    \begin{subfigure}{0.155\textwidth}
        \centering
        \includegraphics[width=\linewidth, trim=0 0 400 0, clip]{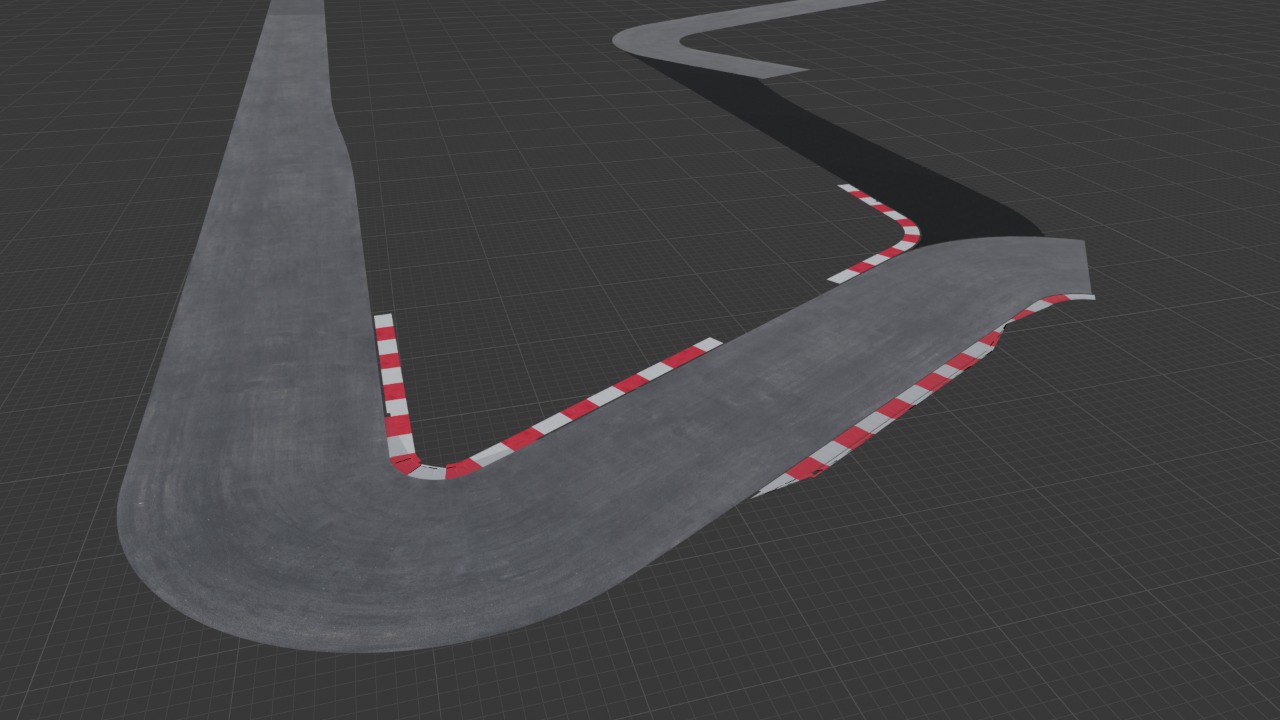}
        \caption{}
        \label{fig:sub2_crg}
    \end{subfigure}
    \hfill
    \begin{subfigure}{0.155\textwidth}
        \centering
        \includegraphics[width=\linewidth, trim=0 0 400 0, clip]{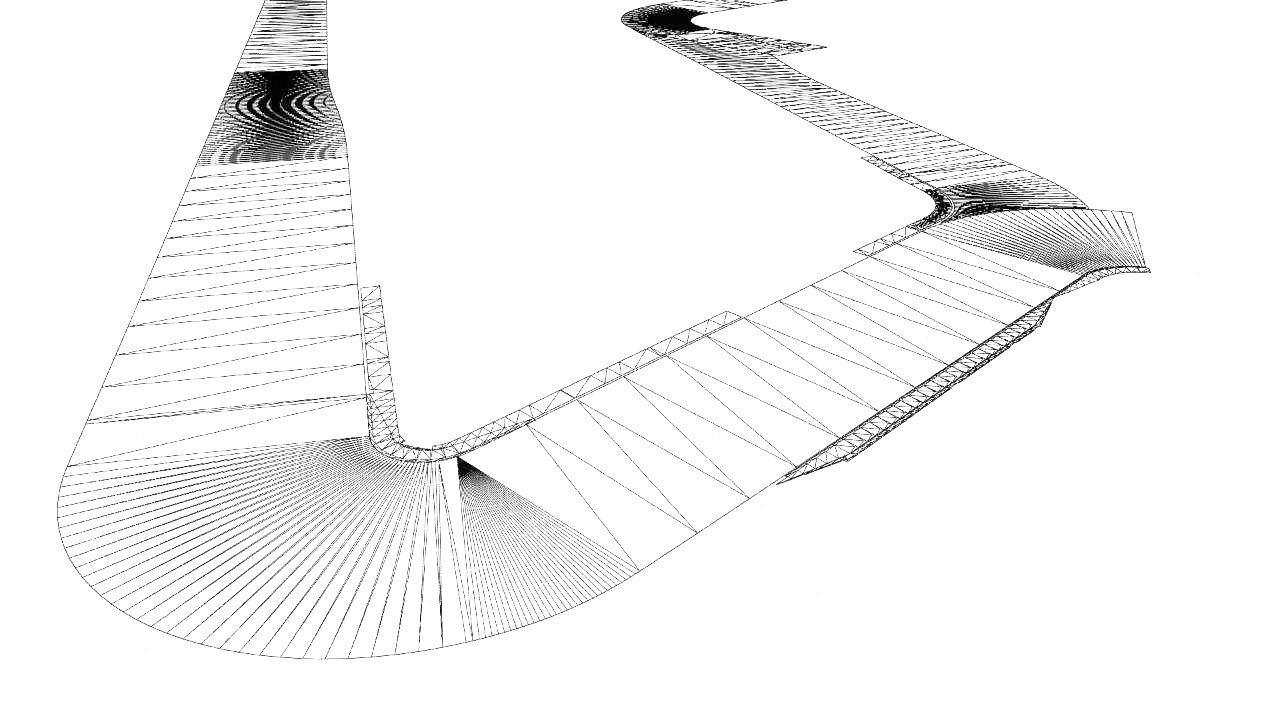}
        \caption{}
        \label{fig:sub3_crg}
    \end{subfigure}
\caption{Race-track exportation process, in a detailed visualization of 
Turn 6 of the Yas Marina circuit (North layout).
}
    \label{fig:vy_filter}
\end{figure}

\subsection{Road Model}\label{subsec:road_model}

The road surface was generated from an existing track file exported 
from standard simulation platforms. 
This preliminary model, shown in Figure~\ref{fig:sub1_crg}, serves as a first 
approximation of the track geometry.
The track representation can subsequently be refined using a 
three-dimensional map reconstructed from the autonomous vehicle’s 
LiDAR measurements. This refinement step is carried out whenever 
significant discrepancies are observed between simulated and measured 
quantities.

This two-stage approach is particularly effective when preparing for 
competitions held on new tracks for which no prior experimental data 
is available. 
The overall workflow is detailed below:

\begin{enumerate}

\item \textbf{Mesh preprocessing.}  
The raw track mesh is edited using common 3D modeling software 
(e.g., Blender~\cite{Blender}) to remove non-road elements such as trees, grandstands, 
and decorative objects. This step isolates a clean road surface suitable 
for numerical processing (shown in Figure~\ref{fig:sub2_crg}).

\item \textbf{STL export and CRG conversion.}  
The processed road geometry is exported in STL (Stereolithography) format, a triangular-facet mesh suitable for numerical processing (shown in Figure ~\ref{fig:sub3_crg}).  
The geometry is converted into the ASAM OpenCRG format, which represents the road surface through
(i) a reference line (track centerline) and
(ii) a regular elevation grid (height map), defined over longitudinal ($u$) and lateral ($v$) coordinates.

\item \textbf{Grid generation and elevation sampling.}  
The elevation grid is generated using user-defined discretization steps 
along $u$ and $v$.  
For each longitudinal position $u$, lateral offsets $v$ are generated 
orthogonally to the reference line until the surface is detected.  
The elevation of each grid node is then computed by querying the STL 
surface, thus populating the CRG height matrix.

\item \textbf{Local refinement and smoothing (optional).}  
Local mesh refinement can be enabled to increase resolution in selected 
regions, such as kerbs or localized road irregularities.  
Additionally, a moving-average smoothing filter may be applied along the 
height direction to reduce sharp elevation discontinuities.  

\item \textbf{Coordinate alignment.}  
Final roto-translational transformations are applied to ensure consistency 
between the simulated environment and real global coordinates.

\end{enumerate}

\noindent
The resulting CRG file is fully compatible with Dymola via the 
OpenCRG block available in the libraries. 
This component reconstructs the 3D road surface from the CRG data and 
allows the assignment of either a constant friction coefficient or a 
spatially varying friction map.

\subsection{Execution time and Numerical Stability}\label{sub_sec:exec_time}

Achieving real-time performance while ensuring a stable and consistent 
simulation requires careful selection of the integration method and its 
time step. 
In our framework, the model is integrated using a fixed-step explicit Euler
scheme with a timestep of $h = 0.001\,\text{s}$, which is required to accurately
capture fast tire dynamics. Poles analysis confirms that all system poles have
negative real parts, while also revealing the presence of fast dynamics.
This indicates that the chosen timestep represents the minimum value ensuring
numerical stability. 
Additional tests were performed with a larger timestep
($h = 0.0012\,\text{s}$). 
While high-speed simulations remain stable, low-speed
scenarios exhibit oscillatory behavior.
Despite the improved execution time, this configuration was ultimately discarded.

To further accelerate simulations, Dymola implements in-line 
integration
, a symbolic numerical approach for solving 
differential-algebraic systems (DAEs). This method embeds discretization 
expressions directly into the model using symbolic manipulation, allowing 
numerical integration to be performed 'in-line' with the system equations. 
By reducing computational overhead, this technique significantly reduces 
the CPU time required for integration.


\subsection{Dymola Experiment Setup for FMU Export}

\begin{figure}[b]
  \centering
    \includegraphics[width=.9\linewidth, trim=0 20 0 90, clip]
    {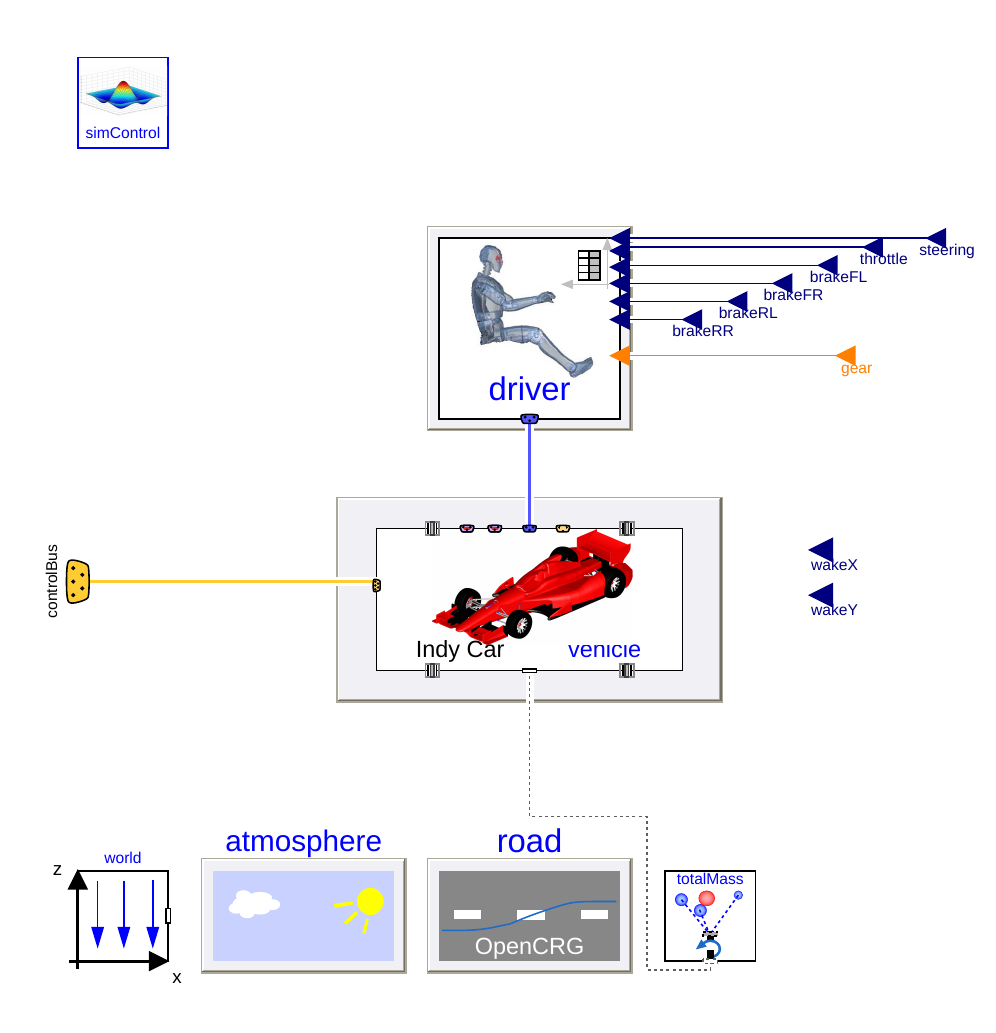}
\caption{Dymola view of the exported FMU experiment, including the vehicle model, open-loop driver, and CRG road blocks. Inputs comprise steering, throttle, four independent braking commands, gear selection, and opponent position (wakeX, wakeY) for slipstream modeling. All relevant output signals are collected in the controlBus.}
\label{fig:fmu_exp}
\end{figure}

A dedicated Dymola experiment, shown in Figure~\ref{fig:fmu_exp}, 
was created specifically for FMU export and integration within the 
external simulation environment. Within this experiment, the vehicle 
model is coupled with the Open-loop driver block and the OpenCRG road block. 

External inputs are provided through Modelica connectors explicitly 
exposed in the FMU interface. These include actuator commands 
and external aerodynamic effects 
supplied by the host simulation environment.


The road block was configured to load a CRG file from a predefined 
directory. Similarly, vehicle parameters and setup configurations are managed 
through external configuration files. A dedicated setup folder contains 
a text file that defines all tunable vehicle parameters, including camber 
and toe angles, suspension characteristics, and 
differential configuration.
The aerodynamic and engine maps are imported from external MAT files, 
whereas the tire models are parameterized through \texttt{.tir} files.

This design ensures flexibility during tuning and validation 
activities, as vehicle setup, parameters, and track configurations 
can be modified without requiring FMU re-export. 

Once exported, the FMU behaves as a pre-compiled black-box model: 
its internal equations are not accessible, and interaction is 
limited to the defined input and output variables exposed through the 
Control-Bus connector. The FMU is exported in version 2.0 and executed 
in \emph{Co-Simulation} mode. In this configuration, it automatically 
advances its internal state using its own solver, ensuring stable and 
efficient integration with the host simulation environment.

\section{Calibration}\label{sec:calib}

The objective of the calibration phase is the unique determination of the micro-parameters characterizing the Pacejka Magic Formula $6.2$. The Pacejka micro-parameters are identified so that the tire model can approximate, with the smallest possible error, the experimentally obtained force and slip values for each vertical load and camber angle condition.

\begin{figure}[t]
    \centering
    \begin{subfigure}[b]{0.235\textwidth}
         \includegraphics[width=\textwidth]{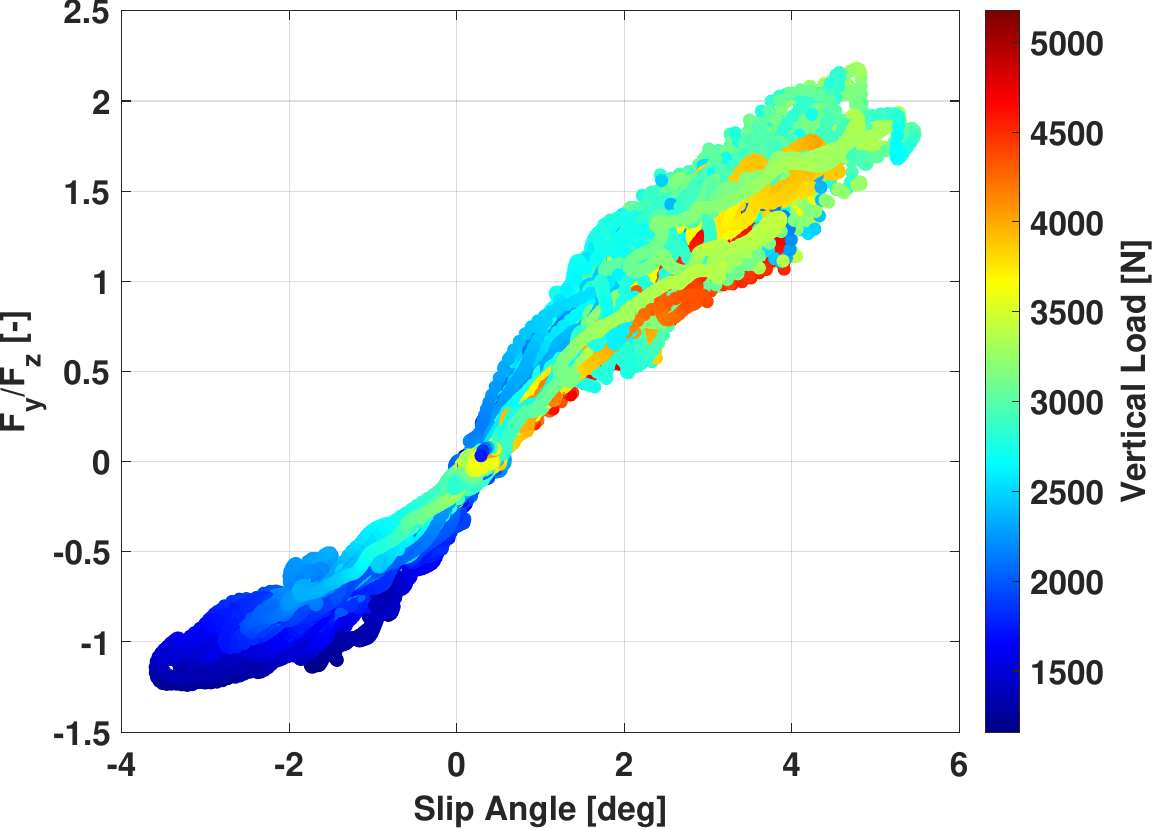}
         \caption{Pure lateral.}
     \end{subfigure}
     \hfill
     \begin{subfigure}[b]{0.235\textwidth}
         \includegraphics[width=\textwidth]{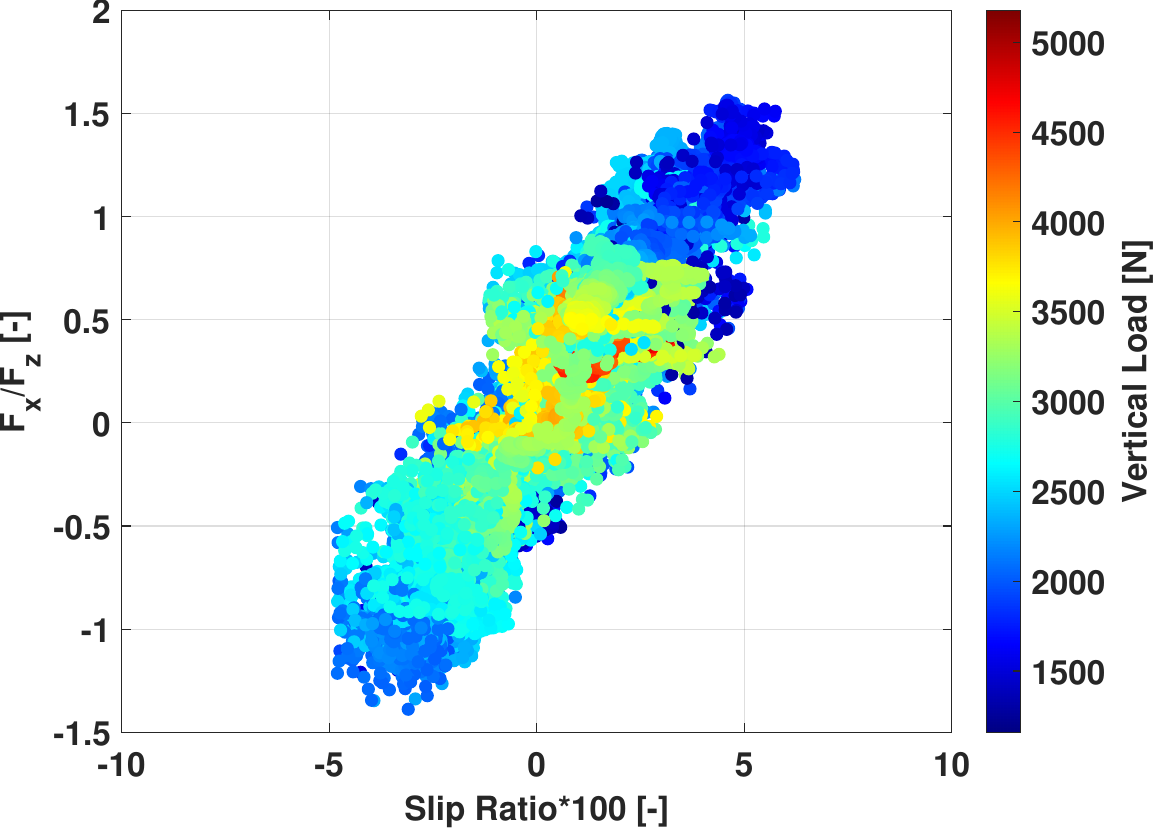}
        \caption{Pure longitudinal.}
     \end{subfigure}
     \caption{T.R.I.C.K outputs for the rear right wheel.}
     \label{fig:trick_outputs}
\end{figure}

The tire–road interaction is characterized through T.R.I.C.K. (Tire/Road Interaction Characterization $\&$ Knowledge)\cite{farroni_trick}, a mathematical model that takes as input a vehicle parameterization and signals acquired from the vehicle CAN bus and provides as output “virtual telemetry” channels containing estimates of forces and slip, as shown in Figures \ref{fig:trick_outputs}.

Through parameterization, the vehicle characteristics can be described numerically. The model includes both more direct parameters (such as mass and wheelbase) and parameters describing more complex vehicle features (such as the aero map and dynamic variations of toe and camber angles). The signals provided as input to T.R.I.C.K. are:
\begin{itemize}
    \item\textbf{Acceleration components}: obtained from Inertial Measurement Units (IMU). These are essential for load transfer estimation.
    \item\textbf{Yaw rate}: obtained from the IMU. It is required for the calculation of lateral forces and wheel slip.
    \item \textbf{Steering angle}: measured by the steering potentiometer. Through a dedicated function, the wheel angle is derived from the steering angle, accounting for kinematic variations in the Ackermann percentage.
    \item \textbf{Wheel rotational speeds}: measured by a toothed wheel sensor on each wheel. These are necessary for slip ratio estimation.
    \item\textbf{Vehicle slip angle}: obtained via Kistler SF-Motion optical sensor. It is required for slip estimation. 
\end{itemize}
Once the tire–road interaction characterization phase is completed, the actual tire model calibration can be performed through an optimization algorithm capable of selecting the best solution among the available candidates\cite{farroni_trick_2}. The algorithm varies the Pacejka micro-parameters within a prescribed range to minimize an objective function representing the deviation between the experimental force values and those computed through the Magic Formula.

The experimental characterization phase and the calibration phase are strictly interconnected. Experimental tests aim to acquire the minimum required number of force-slip data points for each tire operating condition, to simplify and improve the optimization process carried out by the algorithm.

 \leavevmode

\section{Simulation Framework}\label{sec:sim_env}

\subsection{C++ environment for co-simulation}

The FMU emulates the autonomous vehicle during simulation, enabling the testing
of all vehicle-dynamics-related modules, such as localization, planning, and
control, within a closed-loop framework.
Since the entire autonomous driving stack is implemented in C++, 
we decided to integrate the exported vehicle model through the FMI4cpp library~\cite{FMI4cpp}, which provides an interface for FMU-based simulation.

\begin{figure}[!t]
\centering
\resizebox{\columnwidth}{!}{%
\begin{tikzpicture}[
    urblock/.style={rectangle, draw, fill=gray!20, text width=7em, text centered, rounded corners, minimum height=3em},
    block/.style={rectangle, draw, fill=blue!20, text width=5em, text centered, rounded corners, minimum height=3em},
    fmublock/.style={rectangle, draw, fill=red!20, text width=5em, text centered, rounded corners, minimum height=3em},
    deviceblock/.style={rectangle, draw, fill=green!20, text width=5em, text centered, rounded corners, minimum height=3em},
    arrow/.style={->, thick, >=stealth}
]

\node[block] (simnode) {Sim-node};
\node[fmublock, below=1.75cm of simnode] (fmumgr) {FMU-manager};
\node[fmublock, left=2.5cm of fmumgr] (fmu) {FMU};
\node[deviceblock, right=2.5cm of fmumgr] (device) {Device};
\node[urblock, above=1.25cm of simnode] (autonomous) {Autonomous Driving Stack};
\node[above=1.2cm of fmu] (config) {\texttt{config.txt}};

\draw[arrow] (autonomous) -- (simnode) node[midway, left]{(\textbf{U, w})};
\draw[arrow] (simnode) -- (fmumgr) 
    node[midway, left, align=right]{\textbf{U} \\ \emph{step request}};
\draw[arrow] (fmumgr) -- (fmu) node[midway, above]{\textbf{in / out}};
\draw[arrow] (fmu) -- (fmumgr) node[midway, above]{};
\draw[arrow] (fmumgr) -- (device) node[midway, above]{\textbf{W}};
\draw[arrow] (device) -- (simnode) node[midway, right]{\textbf{w}};
\draw[arrow] (simnode.north) -- ++(0,1) -| (autonomous.south) node[midway, right]{};
\draw[arrow] (config) -- (fmu);

\end{tikzpicture}
}
\caption{Block diagram of the closed-loop C++ simulator. 
In the scheme, \textbf{U} denotes control commands from the autonomous driving stack, 
\textbf{W} is the vehicle state computed by the FMU,
and \textbf{w} represents simulated sensor measurements fed back to the software stack.}
\label{fig:simulator_block}
\end{figure}
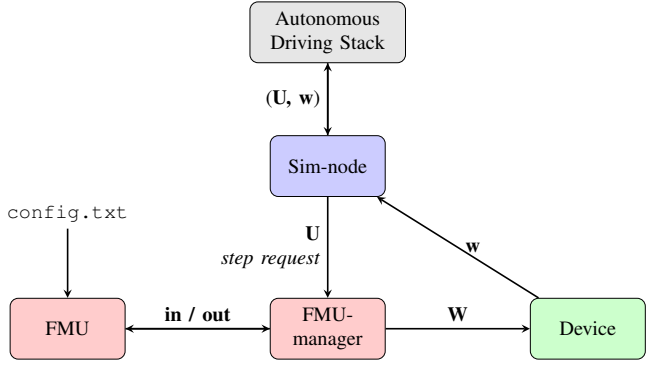

The simulator architecture is organized into three main modules: \emph{Sim-node},
\emph{Fmu-manager}, and \emph{Device}, as shown in Fig.~\ref{fig:simulator_block}. 
Interaction with the FMU is entirely encapsulated within the \emph{Fmu-manager}, 
which wraps the FMI4cpp library and represents the sole interface between 
the simulator and the FMU, ensuring a clear separation of concerns and improved 
maintainability.
Runtime operation follows a fixed-frequency loop (1 ms), coordinated 
by \emph{Sim-node}. At each iteration, it receives control commands from the 
autonomous driving stack and delegates the simulation step to the \emph{Fmu-manager}, 
which handles all FMU operations, including initialization, input updates, time 
stepping, output retrieval, and termination:

\begin{enumerate}

    \item \textbf{Input update:} 
    Vehicle control commands are processed (normalization, steering and brake 
    model application) and written to the FMU.

    \item \textbf{Time integration:} 
    The FMU advances its internal state using its own solver. The simulator can 
    optionally apply a speed-up factor for faster-than-real-time execution
    (particularly useful during automatic simulations \cite{lambertini}). %
    The inputs remain constant during each step.

    \item \textbf{Output retrieval:} 
    Vehicle state variables 
    are read from the FMU, with unit conversions applied to match simulator conventions.

\end{enumerate}

The \emph{Device} module synchronizes with the vehicle state from the FMU
outputs and produces simulated sensor measurements. Each device applies sensor-specific processing, 
including signal normalization, noise injection, and enforcement of sensor-specific 
update rates. Sensor characteristics and parameters are defined through configuration 
files, allowing easy adaptation when switching between different FMUs.

Finally, processed sensor data are published in the same format as real vehicle 
sensors, providing a coherent and realistic vehicle feedback.

\subsection{Graphical User Interface (GUI)}

\begin{figure}[t]
  \centering
    \includegraphics[width=1.\linewidth, trim=2 0 10 0, clip]
    {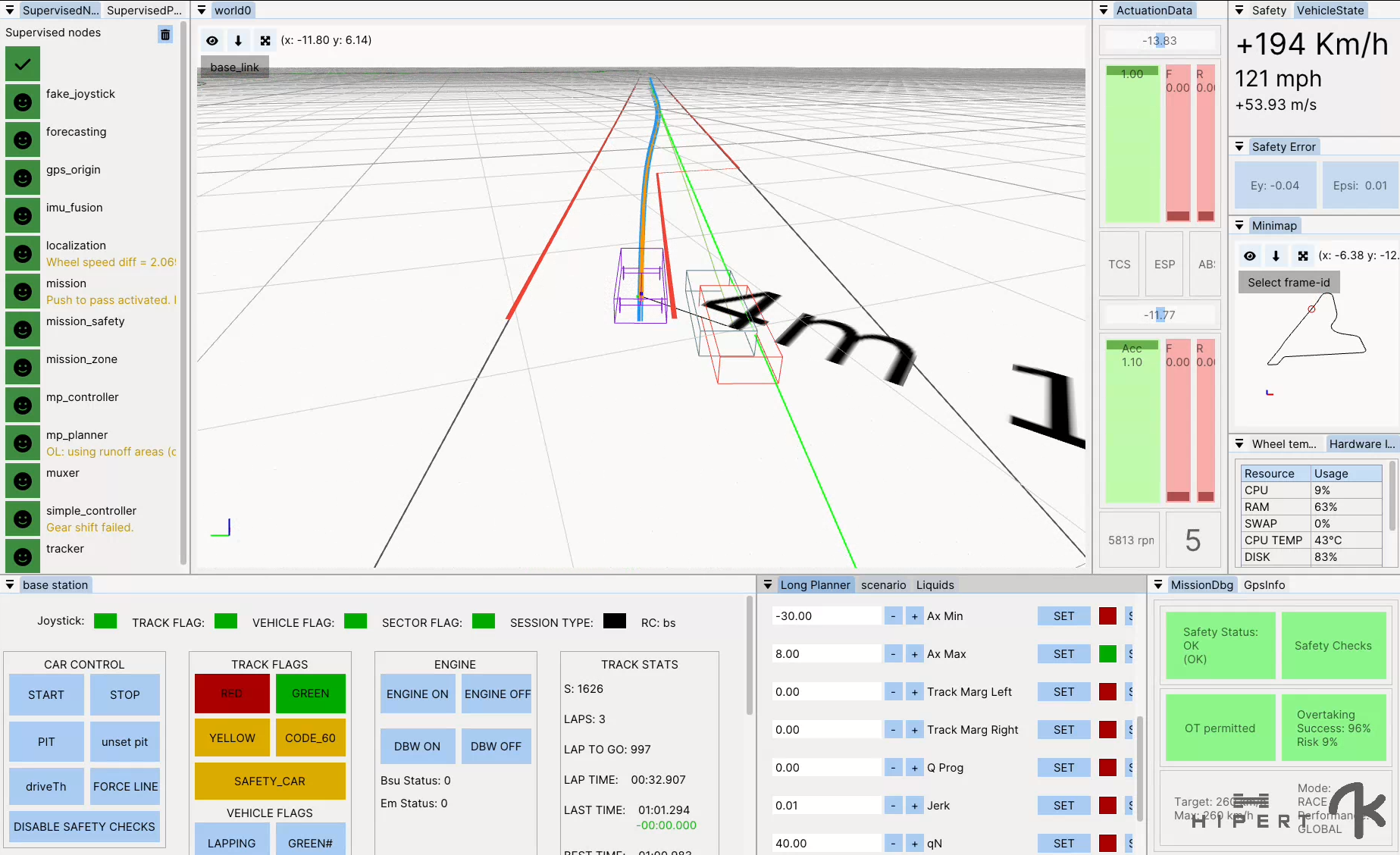}
\caption{The Graphical User Interface (GUI) developed to interact with the simulation environment.}
\label{GUI}
\end{figure}

The simulator can run in both graphical (Figure \ref{GUI}) and headless modes.
The graphical visualization is a proprietary implementation based on the ImGui library~\cite{imgui},
providing a basic 3D wireframe representation of the vehicle.
Additionally, visual cues indicate drivable area boundaries,
planned trajectories, and sensor data. The interface also presents
a comprehensive set of metrics, including speed, steering, throttle,
brake, and gear (for both commands and feedback).

Beyond the visualization of automated, preconfigured simulations,
the interface enables user-friendly interaction with the simulated model.
It provides quick commands for basic vehicle operations, such as starting,
stopping, setting a target speed or changing the reference racing-line. 
Through custom plugins, the interface can dynamically apply parameters that influence the autonomous agent’s driving style at runtime.

Alongside the primary interface, the open-source software PlotJuggler~\cite{pj} visualizes real-time virtual telemetry.


\section{Results}\label{sec:results}

In this section we present the results of the proposed solution by comparing simulated and experimental data.
Specifically, the platform is the Dallara EAV-25 used in the 
Abu Dhabi Autonomous Racing League (A2RL\footnote{\url{https://a2rl.io/autonomous-car-race}})
in the Yas Marina track (North layout).

\subsection{Model Accuracy}

\subsubsection{Tire-road interaction}
\begin{figure}[b]
  \centering
    \includegraphics[width=.94\linewidth, trim=0 0 0 0, clip]
    {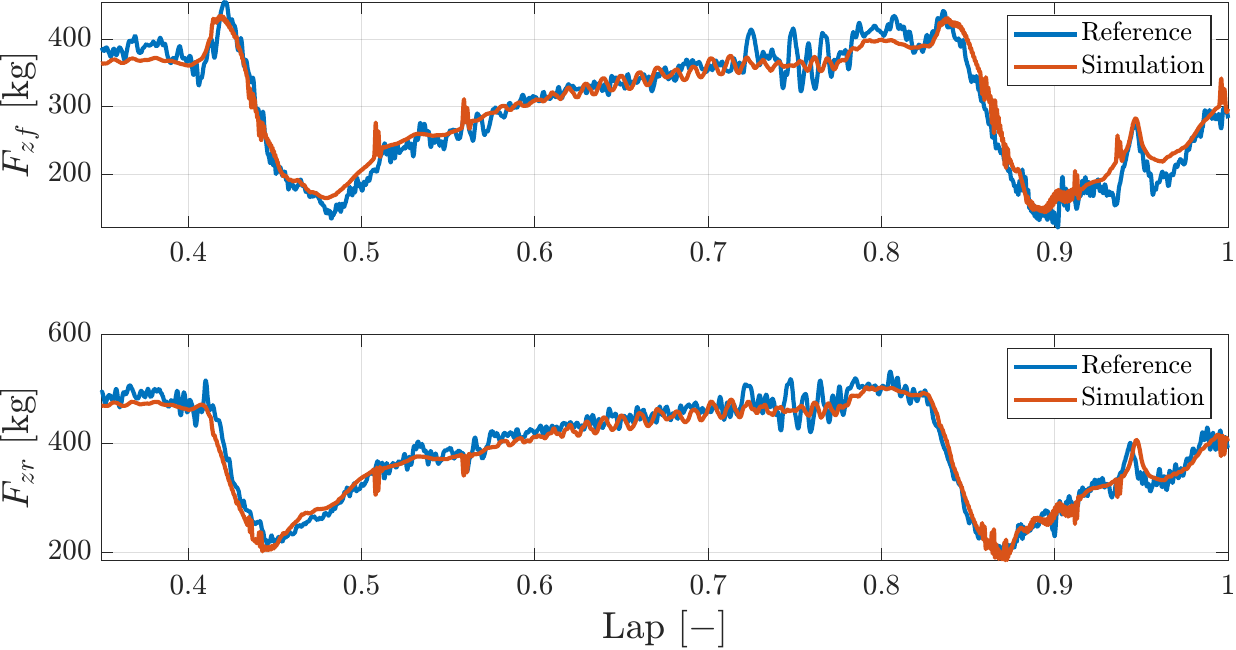}
\caption{Vertical load at the front ($F_{zf}$) and rear ($F_{zr}$) axle over normalized lap distance.}
\label{fig:Fz_per_axle}
\end{figure}

Figure~\ref{fig:Fz_per_axle} compares the measured and simulated wheel 
loads on the second part of the 
racetrack, which includes the most uneven sections. 

High-frequency road irregularities are filtered through the smoothing 
process ($Lap \approx 0.75$). Although this may slightly reduce simulation realism, it is particularly important when a simplified tire–road contact model (e.g., single-point formulations) is adopted, as it ensures a smoother vertical tire dynamic response. 
However, the main features of the track are preserved, including 
low-frequency bumps in the braking zone ($Lap \approx 0.8$) and the 
elevation change / banking in Turn 8 ($Lap \approx 0.95$), which are adequately captured by the model.

\subsubsection{Brake thermal model}
\begin{figure}[t]
  \centering
    \includegraphics[width=1.\linewidth, trim=0 0 110 0, clip]
    {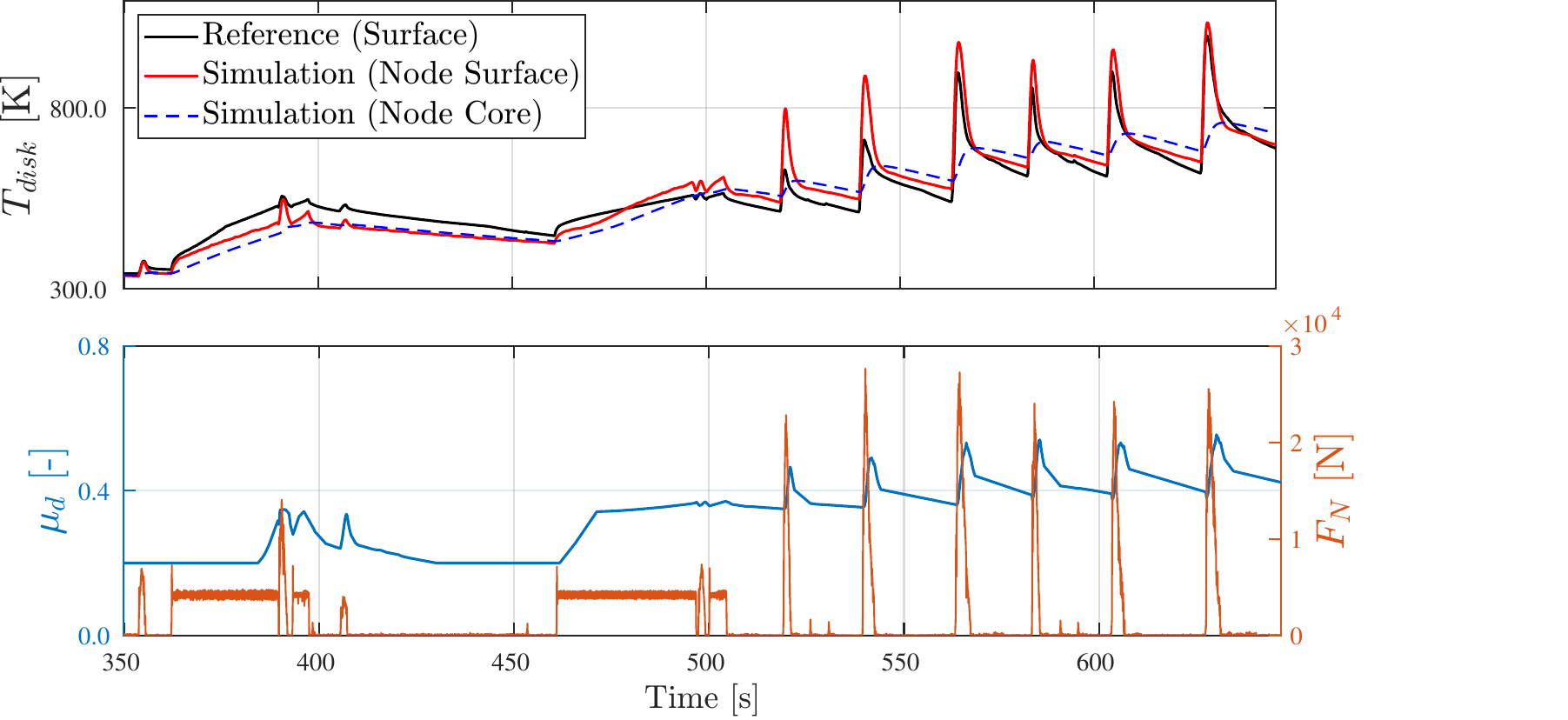}
\caption{Brake thermal model.
Top: simulated surface and core temperatures, with measured surface temperature.
Bottom: friction coefficient ($\mu_d$) from the friction map and applied normal force ($F_N$) at the brake disk.}
\label{fig:compare_brake_results}
\end{figure}

Figure~\ref{fig:compare_brake_results} reports the brake thermal model results for the front-left wheel.

The vehicle starts from standstill with a reduced pad-disk friction coefficient ($\mu_d$) due to cold brake disk. A dedicated warm-up maneuver is then performed, consisting of constant brake inputs ($Time \approx 360$ s and $Time \approx 460$ s), allowing the system to heat up and increase braking performance, as reflected by the friction coefficient.

During the first two braking events ($Time \approx 500$–$550$ s), the surface temperature is significantly overestimated. This is mainly attributed to the assumption of constant thermal conductance between nodes that cannot adequately describe the heat flux in the early thermal transient from an initially cold state. However, after the system reaches its operating temperature range, the thermal model provides a consistent approximation, accurately reflecting the experimental data.

\subsection{Control-related Results}

\subsubsection{Fastest lap analysis}
\begin{figure}[t]
  \centering
    \includegraphics[width=1.\linewidth, trim=0 0 0 0, clip]
    {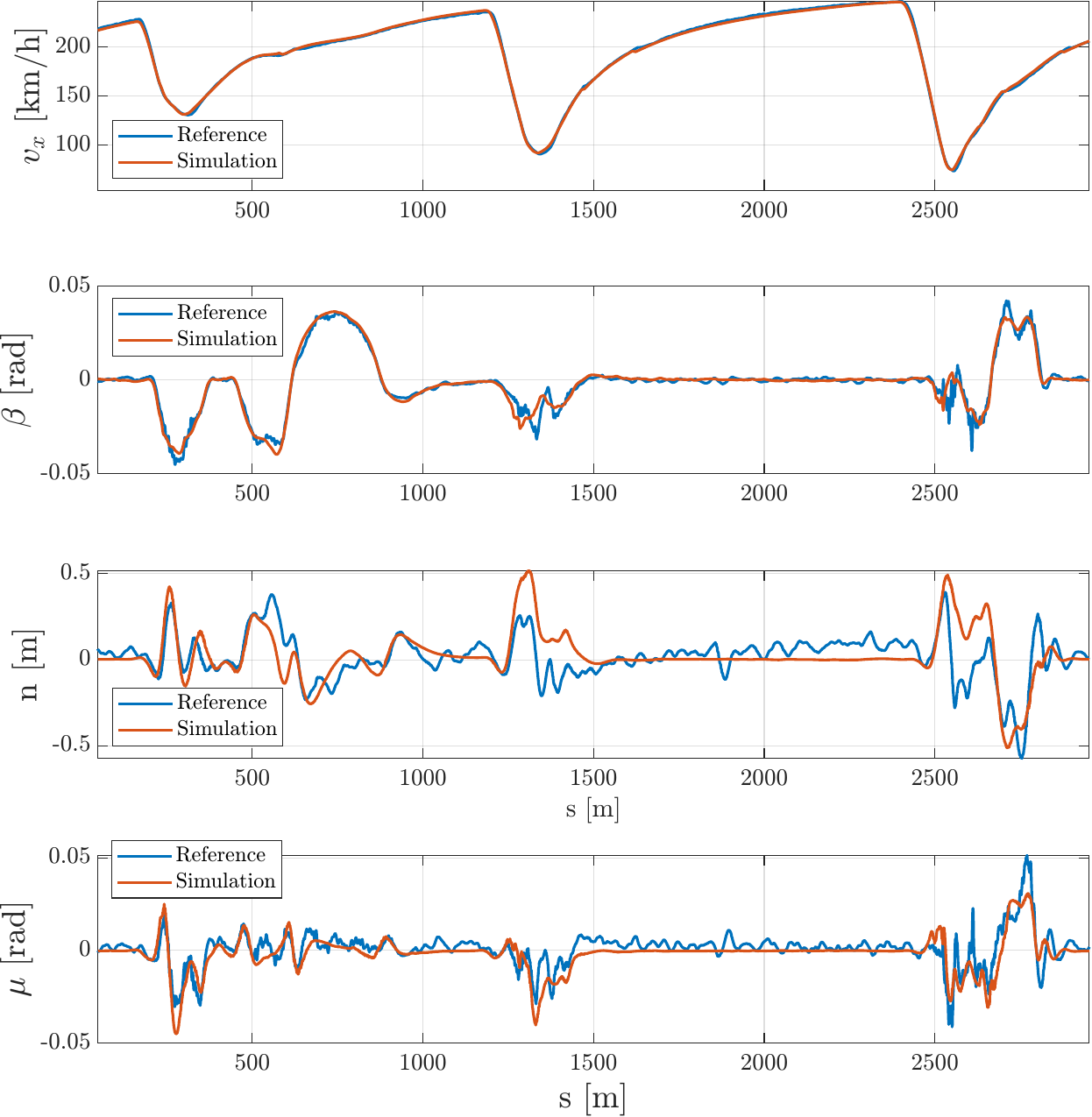}
\caption{Full lap comparison. From top to bottom: longitudinal velocity ($v_x$), sideslip angle ($\beta$), lateral deviation (n), and heading error ($\mu$), plotted 
against lap distance (s).}
\label{fig:full_lap}
\end{figure}

Figure~\ref{fig:full_lap} compares the simulated best lap with the corresponding real-world lap, with the software achieving a lap time within 2\% of the best lap recorded by a former Formula 1 driver using the same car.

The reference lap time is 58.763 s, while the simulated lap time is 58.829 s. Note that lap time is an aggregate metric: the real vehicle features a more powerful powertrain, but also a larger torque delivery (due to modeling simplifications). Therefore, it should not be interpreted as a direct indicator of perfect model matching.

Both longitudinal velocity ($v_x$) and sideslip angle ($\beta$) are accurately replicated, with 
minor discrepancies observed in Turn 5 ($s \approx 1400$ m). 
For autonomous racing applications, the ability to reproduce controller tracking 
errors in simulation is crucial and particularly challenging, 
as centimeter-level accuracy is required. Despite the high fidelity of the 
vehicle model, small discrepancies in the motion field propagate into slight 
mismatches in the simulated lateral error (n), especially in Turn 5 
and Turn 6 ($s \approx 2600$ m).
Overall, the results show good agreement. Quantitative error metrics are reported in Table~\ref{tab:error_metrics}.

\begin{table}[t!]
    \centering
    \caption{Simulation error metrics
    with respect to experimental telemetry.}
    \label{tab:error_metrics}
    \begin{tabular}{lcccc}
        \toprule
         & $v_x$ [m/s] & $\beta$ [rad] & n [m] & $\mu$ [rad] \\
        \midrule
        MAX  & 1.30  & 0.0240  & 0.599 & 0.0348 \\
        RMSE & 0.46  & 0.0035  & 0.143 & 0.0070 \\
        \bottomrule
    \end{tabular}
\end{table}

The simulator was useful to study in advance some planning and control behavior. For example, positive turn-in peak errors are observed in Turn 1 ($s \approx 250$ m), Turn 5, and Turn 6 during the simulation phase. These originate from an overestimation of the tire combined-slip effect within the controller, leading to an anticipatory steering action to compensate for the predicted reduction in lateral force. The same peak errors are then observed in the real car.
Similarly, an oscillatory behavior during the turn-out phase can be observed,
especially in Turn 1 and Turn 8 ($s \approx 2800$ m).
It is worth noting that real-world experiments exhibit larger oscillations due to sensing delays and localization noise.

\begin{figure}[t]
  \centering
    \includegraphics[width=1.\linewidth, trim=0 0 0 0, clip]
    {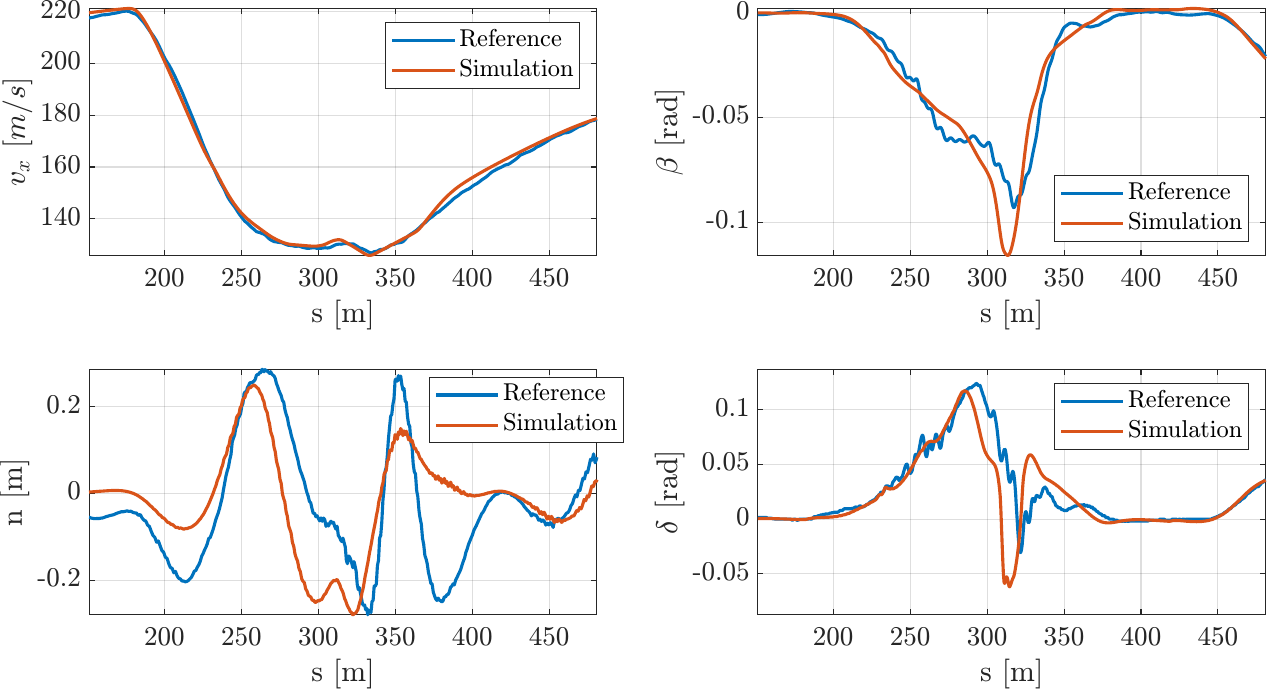}
\caption{Turn~1 detail: longitudinal velocity ($v_x$), sideslip angle 
($\beta$), lateral deviation (n), and steering angle ($\delta$) plotted 
against lap distance (s).}
\label{fig:turn_1}
\end{figure}

\subsubsection{High-dynamics scenario}
Figure \ref{fig:turn_1} shows Turn 1 of the racetrack, where the vehicle frequently 
experiences oversteer events. This behavior is evident in the increase in sideslip angle ($\beta$) and the rapid countersteering action ($\delta$) commanded by the controller. The oversteer is triggered by combined slip at the rear tires, and is consistent with the increase in longitudinal velocity. A close agreement is also observed in the lateral deviation from the planned trajectory.

In simulation, the oversteer is slightly anticipated. This is 
mainly due to the slower engine response observed in the real vehicle, 
caused by turbo-lag. The effect can also be observed in the 
longitudinal velocity, which increases slightly earlier in the 
simulation compared to the experimental data. 
Moreover, the value of $\beta$ is slightly more pronounced, as it occurs under 
higher lateral acceleration, again due to the anticipated traction phase, which 
coincides with the region of maximum curvature.
Despite minor discrepancies in numerical values, the simulator 
accurately reproduces the qualitative dynamic behavior observed in the 
experiment. This provides sufficient confidence to deploy the 
autonomous stack in such critical scenarios and to operate safely
near the grip limit.


\subsection{Real-time performance}

\begin{figure}[t]
  \centering
    \includegraphics[width=1.\linewidth, trim=0 0 0 0, clip]
    {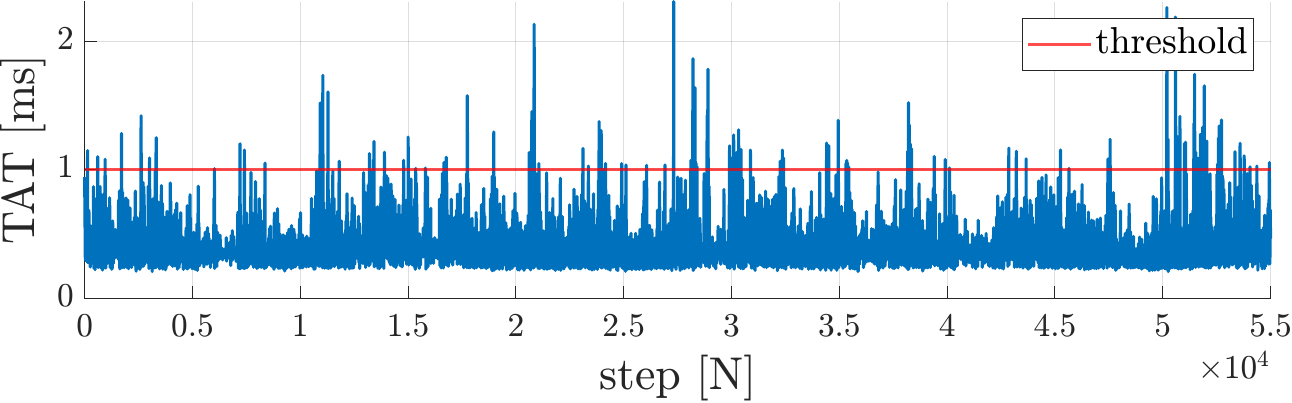}
\caption{Turnaround time of the FMU simulation. The average execution time is               $0.357\,\mathrm{ms}$, while the red line marks the real-time threshold of 
        $1\,\mathrm{ms}$.}
\label{fig:fmu_time}
\end{figure}

Figure~\ref{fig:fmu_time} shows the measured Turnaround time (TAT) while running the full autonomous driving software (including planning, control, and localization) on the same PC.
The FMU is capable of real-time performance, with all observed runtimes consistently below the execution target of $1~\text{ms}$. Only a very small number of peaks slightly exceed this target, and these have no significant 
impact on the model's response.

The simulations were conducted on an Ubuntu 24 system featuring an Intel Core i9-14900HX CPU (32 threads) and 32\,GB of RAM.

\section{Conclusion and Future Work}
\label{sec:conclusion}

The proposed framework provides a vehicle-dynamics-oriented tool for testing Autonomous Racing algorithms, addressing a gap in the literature where vehicle dynamics are often simplified in favor of highly realistic sensor modeling.
The framework demonstrates real-time performance and accurate reproduction of vehicle behavior across both standard and near-limit scenarios, supporting the development of planning and control strategies.
Some aspects of dynamic accuracy require further investigation, including a more detailed powertrain model to better capture turbocharger effects and the extension of the current single-point tire–road contact formulation to a multi-point model, allowing a more accurate description of load when driving over curbs.

Further development will focus on: integrating exteroceptive sensor-simulation to create a unified environment for full-stack validation; the development of a game-oriented graphical interface to collect and compare human driving data; and the exploitation of the FMU interface for the development of data-driven and neural-network-based planning and control approaches \cite{train_fmu_gym}.
Thanks to its reduced sim-to-real gap, the platform also represents a step toward extensive training of AI agents and their real-world deployment.

\bibliographystyle{IEEEtran}
\bibliography{bibliography}

\end{document}